\RequirePackage[svgnames,table]{xcolor}

\documentclass[11pt,letterpaper,logo]{yalearxiv}

\usepackage{graphicx}
\usepackage{float,epstopdf}
\usepackage{bbm}

\usepackage{microtype}

\usepackage{natbib}
\setcitestyle{square}

\usepackage{subcaption}
\usepackage{booktabs}

\usepackage{amsmath}
\usepackage{amssymb}
\usepackage{mathtools}
\usepackage{amsthm}
\usepackage{dsfont}
\usepackage{multicol}
\usepackage{makecell}
\usepackage{multirow} 
\usepackage{amsfonts} 
\usepackage{mathrsfs}
\usepackage[amssymb, thickqspace]{SIunits}
\usepackage{enumitem}
\usepackage{pgfplotstable}
\usepackage{lipsum}		

\usepackage{microtype}
\usepackage{graphicx}
\usepackage{booktabs} 
\usepackage[table]{xcolor}
\usepackage{arydshln}
\usepackage[normalem]{ulem} 

\usepackage{cases}
\usepackage{wrapfig}

\usepackage{url}

\usepackage{thmtools}
\usepackage{thm-restate}
\usepackage{tabu}

\definecolor{huskypurple}{HTML}{4B2E83}

\usepackage{titletoc}

\usepackage{listings}
\lstdefinestyle{promptstyle}{
  basicstyle=\ttfamily\footnotesize,
  breaklines=true,
  breakautoindent=false,
  breakindent=0pt,
  postbreak=\mbox{\textcolor{gray}{$\hookrightarrow$}\space},
  columns=fullflexible,
  keepspaces=true,
  frame=single,
  framesep=5pt,
  xleftmargin=6pt,
  xrightmargin=6pt,
  aboveskip=8pt,
  belowskip=8pt,
  showstringspaces=false,
}

\usepackage{algorithm}
\usepackage{algpseudocode}   
\makeatletter
\def\munderbar#1{\underline{\sbox\tw@{$#1$}\dp\tw@\z@\box\tw@}}
\makeatother

\AddToHook{cmd/appendix/before}{%
  \setcounter{axiom}{0}%
}

\newcommand{\be}{\begin{equation}}
\newcommand{\ee}{\end{equation}}
\newcommand{\bea}{\begin{equation*}\begin{aligned}}
\newcommand{\eea}{\end{aligned}\end{equation*}}

\newtcolorbox{simpleElegantQuote}{
    colback=AliceBlue!50!White,
    colframe=RoyalBlue!75!Black,
    boxrule=0.5pt,
    arc=2mm,
    boxsep=4pt,
    left=10pt, right=10pt,
    top=8pt, bottom=8pt,
    fontupper=\itshape,
}

\title{Self-Correcting Long-Horizon Search Agents\\ via Tree-Structured Memory}

\runningtitle{Self-Correcting Long-Horizon Search Agents via Tree-Structured Memory}

\usepackage{fontawesome5}   

\definecolor{yaleblue}{RGB}{0,58,112}

\author{
Aijun Yang, Qianxue Guo, Ziyi Huang, Yuxuan Chen, Shiyou Qian, and Jian Cao\textsuperscript{*}\\
Shanghai Jiao Tong University \space
\textsuperscript{*}Corresponding Author
}

\hypersetup{colorlinks=true, linkcolor=blue!50!black, citecolor=blue!50!black,
            urlcolor=blue!50!black}

\begin{document}

\begin{abstract}
\vspace{-1mm}
{\centering\section*{Abstract}}
Large language model (LLM)-based search agents answer questions through multi-step interactions with external environments. However, providing complete execution trajectories to the LLM causes unbounded context growth and introduces noise. Existing compression methods reduce context at the cost of important details and often replace erroneous facts without repairing downstream reasoning derived from them. To address this problem, we propose ReTree, a self-correcting tree-structured memory mechanism for search agents. ReTree constructs a bounded per-step reasoning context while preserving source-linked evidence. It models search as an evidence tree whose nodes store bounded summaries, evidence, and revision histories. When newly retrieved evidence contradicts an earlier claim, ReTree traces back to the node where the claim was introduced, replaces outdated evidence, regenerates summaries, prunes affected branches, and resumes search. Source-grounded evidence provenance supports reliable conflict localization and keeps final claims traceable to retrieved passages. Experiments on four public question-answering and search benchmarks show that ReTree consistently outperforms Full-Trajectory ReAct, improving answer accuracy by up to 25.6 percentage points (pp); the average maximum per-step reasoning context of Full-Trajectory ReAct is $1.27$--$1.51\times$ that of ReTree. These results establish ReTree as an effective self-correcting memory abstraction for long-horizon search.
\end{abstract}

\maketitle

\section{Introduction}

Search agents extend language models with iterative retrieval, allowing them to answer questions whose evidence is distributed across documents or unavailable in model parameters. Systems based on interleaved reasoning and action, browser interaction, and retrieval-guided chain-of-thought have made this loop effective for knowledge-intensive and multi-hop questions \citep{yao2023react,nakano2022webgpt,trivedi2023ircot,press2023measuring}. The same loop creates a state-management problem: every query adds observations, tentative conclusions, and source metadata to a trajectory that the policy must read again at the next step.

However, retaining the complete trajectory makes the per-step reasoning context grow with the search horizon. Existing systems control this growth through rolling summaries, reconstructed workspaces, environment-side working memory, or observation masking \citep{wu2025resum,chen2026iterresearch,jiang2026harness,zhang2026masking}. Retrieval itself is also fallible: a search step can introduce irrelevant, stale, or conflicting information. Because later queries and intermediate conclusions are conditioned on the current state, an erroneous premise can steer both what the agent retrieves next and how it interprets subsequent evidence, creating an error cascade across the search trajectory \citep{xu2024conflicts,feng2026tracking}.

These two hazards interact: compression controls context growth but can make an error cascade harder to repair. Repeated synthesis into a compact state may retain a conclusion while obscuring which passage supported it. Even when later evidence corrects the original premise, replacing that fact alone does not invalidate downstream conclusions derived from its former value. A reliable search memory therefore needs more than a short state: it must preserve source-bound evidence traceability and make premise-dependent states identifiable when evidence changes \citep{gao2023alce,gao2023rarr,jung2026meme}.

\begin{figure*}[t]
    \centering
    \includegraphics[width=0.95\textwidth]{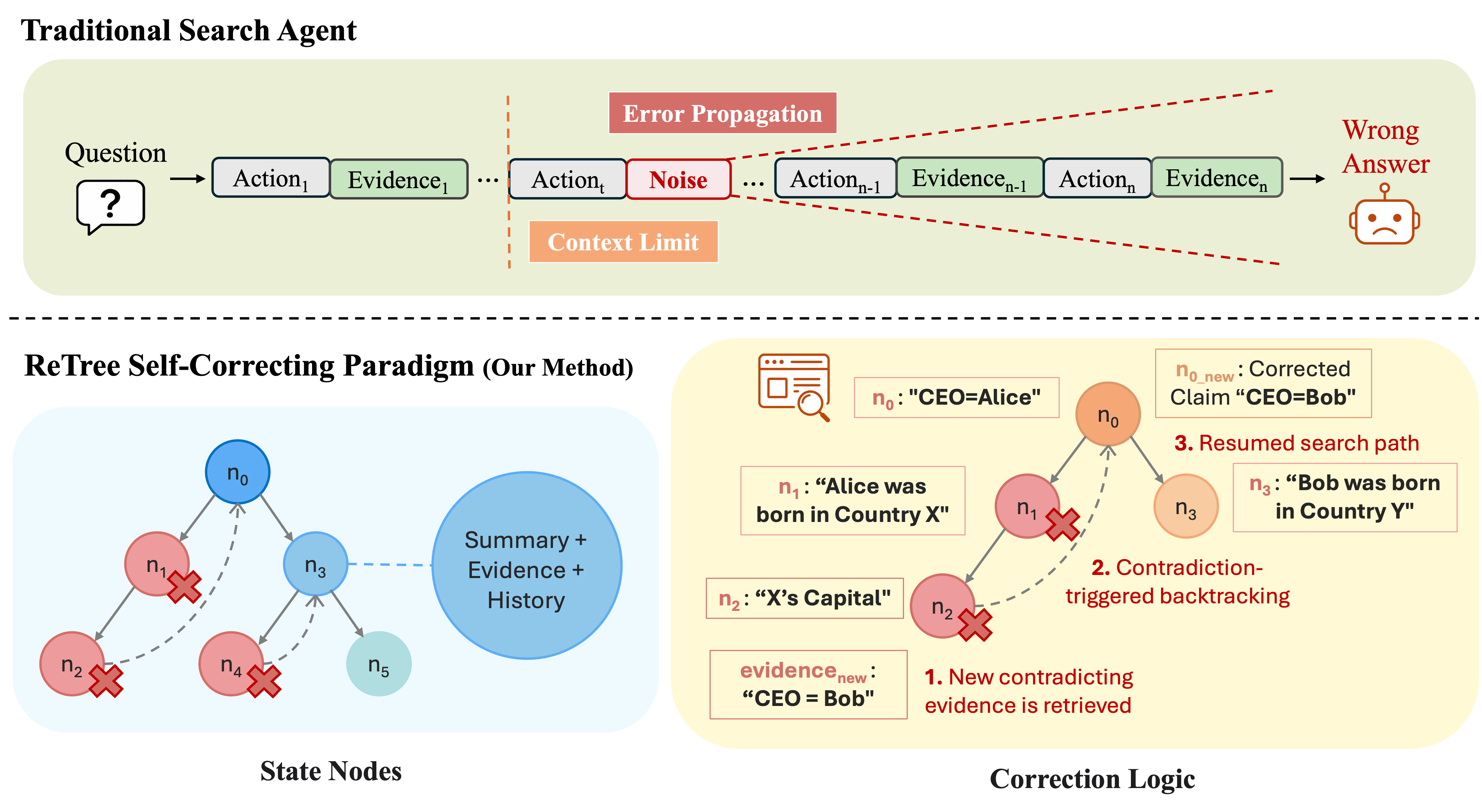}
    \caption{Comparison between full-trajectory agent and ReTree. A full-trajectory agent accumulates observations and propagate an errors (top). ReTree stores source-linked evidence in a revision tree (bottom). When new evidence refutes a claim introduced at $n_0$, ReTree repairs $n_0$, invalidates dependent descendants $n_1$ and $n_2$, and resumes search along the repaired branch through $n_3$.}
    \label{fig:motivation}
\end{figure*}

To repair retrieval-induced error cascades, we introduce \emph{ReTree}, a tree-structured working memory for source-grounded search (Figure~\ref{fig:motivation}). Unlike reasoning trees that organize alternative candidate paths, ReTree is a dependency tree over state evolution and revision history: an edge records that a child search state was derived from its parent's active evidence. Each node stores a bounded task-state summary, locally introduced atomic evidence with stable identifiers and source pointers, and a revision history. The policy sees only the current summary and the top-$k$ relevant evidence items, whereas the external tree retains the full active evidence path.

When new evidence arrives, ReTree first tests it against relevant stored facts and proposes a conflict only for incompatible values under the same entity, slot, scope, and time. A history-aware confirmation then guards the destructive update against ambiguity and repeated flips. Once a conflict is confirmed, ReTree locates the node that introduced the refuted fact, replaces the fact and its source, regenerates that node's summary, prunes dependent descendants, and resumes search from the repaired state. This operation instantiates dependency-directed revision from truth-maintenance systems \citep{doyle1979truth} in an LLM search loop.

ReTree thus separates compact per-step context construction, durable evidence provenance, and structural invalidation. The tree's role is not to parallelize candidate exploration, but to preserve where evidence entered the evolving state and which later states depend on it. This distinguishes ReTree from reasoning trees over candidate thoughts \citep{yao2023tree} and memory indexes that optimize retrieval or localized storage updates \citep{gutierrez2024hipporag,chen2026memforest}.

Our contributions are:
\begin{itemize}
    \item We formalize retrieval-induced error cascades as a structural state-repair problem with three joint requirements: bounded per-step reasoning context, claim-to-source provenance, and dependent-state invalidation.
    \item We develop ReTree, which couples bounded summary-plus-evidence context construction with stable evidence identifiers and contradiction-triggered backtracking, summary regeneration, and subtree pruning.
    \item We implement passage-level source binding that survives memory revision, allowing answer claims to be resolved to current evidence objects and raw retrieved passages.
    \item Across 2,149 questions drawn from four public benchmarks, ReTree improves judge accuracy over Full-Trajectory ReAct by 8.3--25.6 points; the average maximum per-step reasoning context of Full-Trajectory ReAct is $1.27$--$1.51\times$ that of ReTree.
\end{itemize}

\section{Related Work}

\subsection{Search Agents and Long-Horizon Context}

Retrieval-augmented generation conditions answers on external documents \citep{lewis2020rag}, while search agents decide iteratively what to retrieve and how to use it. Representative systems interleave reasoning with actions, decompose compositional questions, or organize browser interaction into source-grounded answer or report generation \citep{yao2023react,press2023measuring,trivedi2023ircot,nakano2022webgpt,shao2024storm}. These methods improve evidence acquisition, but their policy state still largely remains a transcript or generated text.

Long-horizon systems increasingly manage this state explicitly through trajectory summaries, reconstructed reports, environment-side working memory, or observation masking \citep{wu2025resum,chen2026iterresearch,jiang2026harness,zhang2026masking}. ReTree shares the goal of a bounded per-step reasoning context, but focuses on what repeated synthesis can discard: per-fact provenance and the location at which a refuted fact entered dependent reasoning.

\subsection{Agent Memory and Tree-Structured State}

Agent memory systems organize information beyond a raw interaction log by retrieving salient observations, managing tiered context, or maintaining structured long-term records \citep{park2023generative,packer2024memgpt,gutierrez2024hipporag,xu2025amem,chhikara2025mem0,yan2025gam}. These systems primarily target recall across interactions rather than repair within a single evolving search trajectory.

Trees also appear in deliberate reasoning and memory indexing \citep{yao2023tree,shinn2023reflexion,chen2026memforest}. ReTree differs in the semantics of its edges and updates: a child is a search state derived from its ancestor's evidence, and a contradiction causes the introducing ancestor to be repaired and its dependent descendants to be invalidated. The structure is therefore a revision-aware evidence lineage rather than a temporal hierarchy or a search tree over candidate thoughts.

\subsection{Attribution, Conflict, and Belief Revision}

Attribution methods require generated claims to be backed by inspectable evidence \citep{gao2023alce,gao2023rarr,asai2024selfrag}. Knowledge-conflict research studies when external context should override stored or parametric beliefs \citep{xu2024conflicts,xie2024adaptive}. ReTree is narrower than a general belief-revision system: it repairs source-grounded evidence during web search. Its backtracking operation follows the truth-maintenance principle that changing a justification should invalidate dependent conclusions \citep{doyle1979truth}, implemented here through ancestry and subtree pruning rather than a general logical dependency graph.

\section{Problem Formulation}

Let $q$ be a question and let $M_t$ denote the agent's external memory after $t$ retrieval-and-update steps, with $M_0$ the initial state. At search step $t\in\{1,\ldots,T\}$, a context-construction function $\mathcal{C}_k$ produces a task-relevant per-step reasoning context from $M_{t-1}$ under evidence budget $k$. Conditioned on this context, policy $\pi_\theta$ selects either a search query or a stop action,
\begin{equation}
\begin{aligned}
a_t&\sim\pi_\theta\!\left(\cdot\mid q,\mathcal{C}_k(M_{t-1},q)\right),\\
a_t&\in\mathcal{A}_{\mathrm{search}}\cup\{\textsc{Stop}\}.
\end{aligned}
\end{equation}
For a search action $a_t$, the environment returns passages $P_t$ and an extractor produces a set of newly acquired, source-bound atomic evidence, denoted by $\Delta E_t$:
\begin{equation}
\begin{aligned}
\Delta E_t&=\{e_j=(j,x_j,u_j)\mid j\in J_t\},\\
M_t&=\mathcal{U}(M_{t-1},\Delta E_t),
\end{aligned}
\end{equation}
where $J_t$ contains the stable identifiers assigned at step $t$, $x_j$ is a self-contained factual statement, $u_j$ is its source URL, and $\mathcal{U}$ is the memory-update operator that incorporates the new evidence $\Delta E_t$ into the previous memory. On stopping, the agent returns an answer $y$ and claims $C=\{c_i\}$ linked to supporting evidence identifiers.

Given this interaction model, a useful search memory must satisfy three requirements:
\begin{enumerate}
    \item \textbf{Bounded context.} The per-step reasoning context $\mathcal{C}_k(M_{t-1},q)$ should remain bounded as $t$ grows.
    \item \textbf{Traceability.} Every answer claim should remain traceable to retrieved passages after repeated updates.
    \item \textbf{Dependency-consistent revision.} Revisions should invalidate dependent state.
\end{enumerate}
To formalize the third requirement, let $\mathcal{Z}(M_{t-1})$ denote the set of all state components stored in memory $M_{t-1}$ before step $t$, and define
\begin{equation}
\operatorname{Dep}_{t-1}(j)
=\{z\in\mathcal{Z}(M_{t-1}):e_j\leadsto z\},
\end{equation}
where $e_j\leadsto z$ denotes direct or transitive derivation dependence.
If new evidence at step $t$ revises $e_j$ to $e'_j$, let $\mathcal{I}_t$ denote the components invalidated by the update and $\mathcal{G}_t(e'_j)$ those regenerated under the corrected evidence. A dependency-consistent update must ensure
\begin{equation}
\operatorname{Dep}_{t-1}(j)
\subseteq\mathcal{I}_t\cup\mathcal{G}_t(e'_j).
\label{eq:dependent-invalidation}
\end{equation}
Equation~\eqref{eq:dependent-invalidation} requires every component that depends on the revised evidence to be either invalidated or regenerated under its corrected value. Traceability can be lost when facts and sources are repeatedly synthesized into an undifferentiated state; dependency consistency is violated when a memory replaces $e_j$ but retains components derived from its former value.

\section{Methodology}

\begin{figure*}[t]
    \centering
    \includegraphics[width=0.93\textwidth]{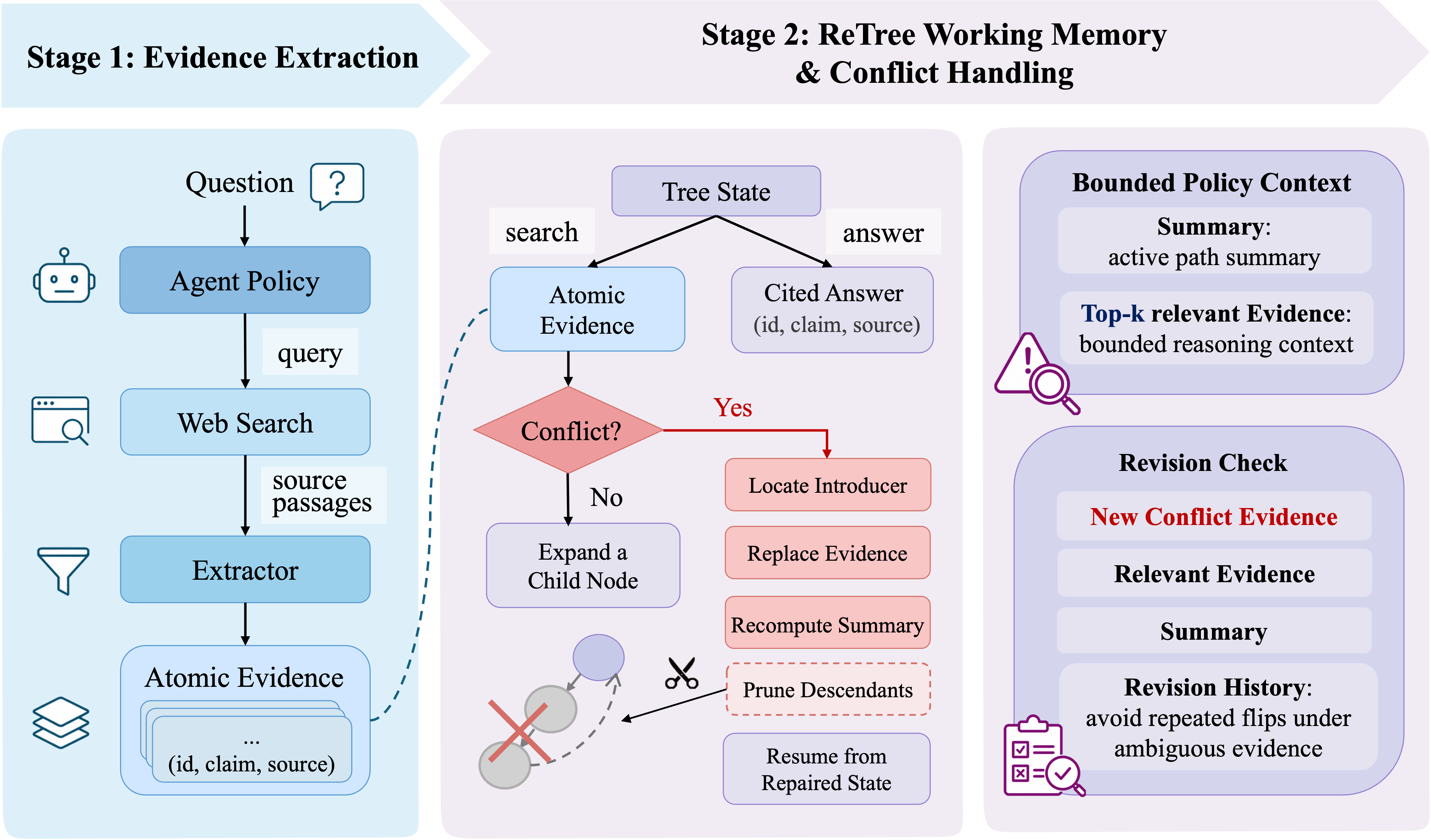}
    \caption{ReTree architecture and LLM contexts. Retrieved passages are converted into source-bound atomic evidence (left). Compatible evidence expands the tree, whereas confirmed conflicts trigger introducer localization, evidence replacement, summary regeneration, descendant pruning, and resumed search (middle). The policy receives a bounded summary-plus-top-$k$ view, while conflict confirmation additionally uses targeted evidence, the branch summary, and revision history (right).}
    \label{fig:overview}
\end{figure*}

Figure~\ref{fig:overview} gives an end-to-end view of evidence extraction, bounded per-step context construction, conflict handling, and source-linked answer generation.

\subsection{Evidence-Tree State}

After $t$ updates, ReTree instantiates the memory as $M_t=(\mathcal{T}_t,n_t)$, where $\mathcal{T}_t$ is an external rooted tree and $n_t$ is its active node. Each node $n\in\mathcal{T}_t$ is
\begin{equation}
n=(s_n,E_n,H_n,p_n),
\end{equation}
where $s_n$ is a compact task-state summary, $E_n$ contains evidence introduced locally at $n$, $H_n$ records revisions, and $p_n$ points to the parent. The accumulated evidence available at $n$ is the union along its root-to-node path,
\begin{equation}
A(n)=\bigcup_{v\in\operatorname{path}(r,n)} E_v.
\end{equation}
Evidence objects are not copied into the summary: their identifiers and URLs remain stable in $A(n)$ and in a run-level evidence map. The summary records the requested answer slot, resolved intermediate slots, remaining open slots, and unresolved candidates in at most 140 words.

The policy never receives all evidence on the active path. ReTree instantiates the context-construction function from the problem formulation as
\begin{equation}
\mathcal{C}_k(M_t,q)
=s_{n_t}\;\Vert\;\operatorname{TopK}_k(A(n_t),q\Vert s_{n_t}),
\end{equation}
where $\Vert$ denotes concatenation, retrieval uses lexical relevance in the current implementation, and $k=5$. With bounded summaries and atomic evidence items, the per-step reasoning context is $O(1)$ in the trajectory length. External storage can still grow with collected evidence; ReTree bounds the per-step reasoning context, not total memory.

\subsection{Expansion and Source Binding}

For search action $a_t$, the environment returns up to five passages $P_t$. The extractor emits at most six question-relevant atomic facts $\Delta E_t$ and selects each fact's source by passage index. Binding by index avoids asking the language model to reproduce a URL, which can silently make an otherwise correct citation unverifiable. If the new facts are compatible with the active path, ReTree creates a child $n_t$ of $n_{t-1}$ whose local evidence is $E_{n_t}=\Delta E_t$ and whose summary compresses the parent state with the new facts. This append-only step preserves where every fact entered the trajectory.

The query policy treats the question as an answer contract. It tracks the requested value type, intermediate slots, and missing relations, and may stop only after at least one real retrieval and after evidence supports the final requested slot. This restriction prevents a seeded or parametric answer from bypassing the search process.

\subsection{Contradiction-Triggered Backtracking}

New facts are compared with the top-$k$ existing facts most relevant to the query and retrieved content. This initial conflict call receives the question, selected existing facts, and new facts, but not the branch summary. A backtrack is allowed only when a new fact gives an incompatible value for the same entity and attribute under the same scope and time; refinements, different entity senses, and competing but compatible candidates expand the tree instead. If this call proposes a backtrack, a second revision call receives the question, branch summary, targeted existing fact, new facts, and the target's change history. It applies the repair only when the contradiction is same-scope and better supported, thereby reducing oscillation between previously rejected values.

Suppose the conflict judge identifies evidence $e_j$ as refuted by $\Delta E_t$. Let $\iota(j)$ denote the node that introduced $e_j$. ReTree replaces the evidence text and source at $\iota(j)$ and appends a record $(x_j^{\mathrm{old}},x_j^{\mathrm{new}},\rho)$ to $H_{\iota(j)}$. It then regenerates $s_{\iota(j)}$ solely from corrected accumulated evidence and removes all descendants of $\iota(j)$. Search resumes by setting $n_t=\iota(j)$, so later steps reconstruct the missing downstream chain from the repaired premise.

Ancestry is a conservative proxy for semantic dependence: a descendant can contain facts that are actually independent of the revised premise. Hard pruning may therefore over-invalidate useful state. ReTree deliberately favors repair safety over state reuse by removing every descendant that could have been conditioned on the refuted evidence; the cost is potentially redundant retrieval and regeneration. Selective invalidation would require explicit evidence-to-summary and evidence-to-query dependency tracking. This trade-off is appropriate for an active search branch but not for tasks that require unrestricted recall of every historical statement.

Algorithm~\ref{alg:retree} summarizes the loop.

\begin{algorithm}[t]
\caption{ReTree Search with Evidence Revision}
\label{alg:retree}
\begin{algorithmic}[1]
\Require question $q$, step budget $T$, evidence budget $k$
\Ensure answer $y$ with evidence-linked citations
\State Initialize tree $\mathcal{T}$ with root $r$ and current node $n \gets r$
\For{$t=1$ to $T$}
    \State $C \gets \operatorname{Summary}(n) \oplus \operatorname{TopK}(A(n),q,k)$
    \State $a \gets \operatorname{Policy}(q,C)$
    \If{$a=\textsc{Stop}$ \textbf{and} $\operatorname{Supported}(A(n),q)$}
        \State \textbf{break}
    \EndIf
    \State $P \gets \operatorname{Retrieve}(a)$; $\Delta E \gets \operatorname{Extract}(P,q)$
    \State $E \gets \operatorname{TopK}(A(n),q \oplus P,k)$
    \State $e \gets \operatorname{ProposeConflict}(q,E,\Delta E)$
    \If{$e \neq \varnothing$ \textbf{and} $\operatorname{ConfirmRevision}(q,n,e,\Delta E,H_{\iota(e)})$}
        \State $u \gets \iota(e)$; replace $e$ at $u$ with supported evidence from $\Delta E$
        \State append revision record to $H_u$; regenerate $\operatorname{Summary}(u)$
        \State prune descendants of $u$; $n \gets u$
    \Else
        \State create child $v$ of $n$ with $\Delta E$ and updated summary; $n \gets v$
    \EndIf
\EndFor
\State \Return $\operatorname{Generate}(q,A(n))$
\end{algorithmic}
\end{algorithm}
\subsection{Claim-Level Attribution}

The final generator decomposes its reasoning into atomic claims and assigns each claim one or more evidence identifiers. A run-level map resolves an identifier to the current evidence text and URL, while a passage map stores the raw text originally returned for each URL. Attribution evaluation therefore follows
\begin{equation}
c_i \rightarrow j \rightarrow (x_j,u_j)
\rightarrow \operatorname{Passage}(u_j),
\end{equation}
and a separate NLI judge labels whether $\operatorname{Passage}(u_j)$ entails, is neutral toward, or contradicts $c_i$. This checks the cited passage rather than the claim in isolation. A report baseline uses the same retrieval backend but retains only a synthesized report and an unordered URL bag, so claim-source links must be reconstructed post hoc. Because this evaluation requires claim extraction and passage-level entailment checks, we use it as a targeted diagnostic rather than a benchmark-wide primary metric.

\subsection{Computation Complexity and Failure Modes}

The current implementation retrieves top-$k$ evidence by lexical overlap in $O(|A(n)|)$ time and locates an evidence introducer by tree traversal in $O(|\mathcal{T}|)$ time. Both can be reduced with vector and identifier indexes; language-model and search calls dominate runtime in our setting. More importantly, correction quality is limited by the conflict judge. A false negative preserves contamination, whereas a false positive can over-prune useful evidence. The same-scope conflict rule and history reconciliation reduce, but do not eliminate, these errors.

\begin{table*}[!t]
\centering
\begingroup
\small
\setlength{\tabcolsep}{4.5pt}
\begin{tabular}{@{}lrccccrr@{}}
\toprule
& & \multicolumn{4}{c}{Answer Quality (Accuracy / EM, \%)}
& \multicolumn{2}{c}{Max Context (chars)} \\
\cmidrule(lr){3-6}\cmidrule(l){7-8}
Dataset & $n$ & ReTree & FlatUpdate & ReportMem. & Full ReAct & ReTree & Full ReAct \\
\midrule
Bamboogle & 125 & \textbf{61.6}/47.2 & 58.4/\textbf{48.0} & 54.4/37.6 & 36.0/28.0 & 1,116 & 1,474 \\
2Wiki & 600 & 50.5/\textbf{31.3} & 45.8/25.8 & \textbf{50.8}/25.8 & 36.5/24.8 & 1,068 & 1,521 \\
HotpotQA & 600 & \textbf{50.5}/32.3 & 48.3/\textbf{32.5} & 48.8/27.3 & 42.2/29.3 & 1,211 & 1,542 \\
FRAMES & 824 & \textbf{31.8}/\textbf{19.5} & 27.9/16.7 & 26.0/13.0 & 15.8/10.1 & 1,274 & 1,920 \\
\midrule
Overall & 2,149 & \textbf{44.0}/\textbf{28.0} & 40.4/25.5 & 40.9/22.0 & 30.1/20.6 & 1,190 & 1,677 \\
\bottomrule
\end{tabular}
\endgroup
\caption{Results on four public benchmarks with Qwen3-8B. Answer quality is reported as judge accuracy / exact match (EM), in percent. Bamboogle and FRAMES use their full evaluation sets; 2Wiki and HotpotQA use fixed 600-question subsets. Avg. Max Policy Context is the mean, over questions, of the largest rendered policy-memory context observed during a run, measured in characters. The Overall row pools all 2,149 questions. The best accuracy and EM are bolded separately.}
\label{tab:main-results}
\end{table*}

\section{Experiments}

We study three questions: whether ReTree improves answer quality over a full-trajectory search agent, whether its structural repair improves on bounded flat-update and report-style memories, and how a bounded per-step policy-memory context affects context growth and aggregate model usage.

\subsection{Datasets}

We evaluate on four public search benchmarks with complementary reasoning demands: Bamboogle \citep{press2023measuring}, HotpotQA \citep{yang2018hotpotqa}, 2WikiMultiHopQA \citep{ho2020twiki}, and FRAMES \citep{krishna2025frames}. Together, these datasets cover a comprehensive spectrum of multi-hop search challenges, including compositional gap exposure, explicit reasoning paths, and realistic fact-seeking queries. Our evaluation suite comprises all 125 Bamboogle questions, all 824 FRAMES questions, and fixed 600-question subsets of 2WikiMultiHopQA and HotpotQA, yielding 2,149 questions per method.

\subsection{Metrics}
We measure performance across three dimensions:

\begin{itemize}
    \item \textbf{Answer Quality:} Judge Accuracy (evaluated via GPT-5 against gold answers under type constraints) and normalized Exact Match (EM).  
    \item \textbf{Provenance}: Citation Precision (CitePrec), Recall (CiteRec), and Uncited Rate (Uncited), verified via NLI-based passage entailment on a fixed 600-question FRAMES subset shared by all evaluated methods.  
    \item \textbf{Context \& Compute}: Avg. Max Per-Step Policy Context (the mean across questions of each question's largest rendered policy-memory context, measured in characters), total model calls, and token usage. This context metric covers the dynamic memory payload supplied to the policy, excluding fixed instructions and other LLM calls. Wall-clock latency is omitted due to live-search network variance.
\end{itemize}

\subsection{Settings}
\begin{itemize}
    \item \textbf{Agent Configurations.} All agents use Qwen3-8B as the backbone policy, the same Google Search API backend, and identical question ordering. Each agent is restricted to at most eight searches per question (returning up to five passages each) and must perform at least one retrieval before answering. Regarding context management, Full-Trajectory ReAct retains the complete, uncompressed observation transcript. ReTree and FlatUpdate bound the context via a 140-word task-state summary alongside the top-5 lexically relevant evidence items, whereas ReportMemory maintains a single evolving report capped at 200 words. All runs use seed 0 with fixed hyperparameters across all benchmarks.
    \item \textbf{Implementation Details.} Local inference uses Qwen3-8B in FP16 through an OpenAI-compatible SGLang endpoint on two NVIDIA A40 (48GB) GPUs, using Python 3.10.20, PyTorch 2.2.1, and CUDA 12.1. Live web search and GPT-5 evaluation use hosted APIs.
\end{itemize}

\subsection{Comparison Approaches}

We select three controls that progressively isolate the contribution of ReTree's memory structure. \textbf{Full-Trajectory ReAct} is the primary baseline: it follows the interleaved reasoning-and-action protocol of ReAct \citep{yao2023react} and retains the complete observation trajectory. It tests whether bounded external memory improves over the standard transcript-based search state.

\textbf{FlatUpdate} is a mechanism-matched control for flat
memory updates, in which individual
records are added or revised in place, as in record-oriented agent memories such as Mem0 \citep{chhikara2025mem0}. It shares ReTree's 140-word task summary, top-5 evidence budget, and conflict detector, but maintains evidence in a flat list. When a conflict is detected, it replaces the refuted fact without locating its introducing state, rebuilding dependent summaries, or pruning downstream reasoning. FlatUpdate cleanly isolates the specific contribution of dependency-directed structural revision.

\textbf{ReportMemory} represents iterative synthesis methods that periodically compress a trajectory or reconstruct an evolving research report, as in ReSum and IterResearch \citep{wu2025resum,chen2026iterresearch}. It folds each retrieval into a single bounded summary while retaining an unordered set of visited URLs, discarding explicit fact-level dependency lineage. ReportMemory tests whether compact context synthesis alone is sufficient for long-horizon search, providing a clear contrast to ReTree's structured, revisable state repair.

\subsection{Results}

\textbf{Answer quality.} Table~\ref{tab:main-results} shows that ReTree improves judge accuracy over Full-Trajectory ReAct on every dataset by 8.3--25.6 percentage points (pp). Pooled over 2,149 questions, the gain is 13.9 pp in judge accuracy and 7.4 pp in EM. Among the stronger bounded controls, ReTree exceeds FlatUpdate by 2.2--4.7 pp on every dataset. It also outperforms ReportMemory on Bamboogle, HotpotQA, and FRAMES by 1.7--7.2 pp, while remaining within 0.3 pp on 2Wiki. Thus, ReTree ranks first on three of four datasets and reaches 44.0\% overall accuracy, 3.0 pp above ReportMemory, the strongest aggregate baseline. These results separate the benefit of structural repair from both unbounded history retention and bounded synthesis alone.

\begin{table}[!htb]
\centering
\begingroup
\small
\renewcommand{\arraystretch}{1.18}
\begin{tabular*}{\columnwidth}{@{\extracolsep{\fill}}lrrrr@{}}
\toprule
& \multicolumn{4}{c}{Claim-level attribution (\%)} \\
\cmidrule(l){2-5}
System & $\mathrm{CitePrec}\uparrow$ & $\mathrm{CiteRec}\uparrow$ & $\mathrm{CiteF1}\uparrow$ & $\mathrm{Uncited}\downarrow$ \\
\midrule
ReTree & \textbf{44.5} & \textbf{41.3} & \textbf{42.8} & 14.8 \\
ReportMem. & 16.8 & 21.4 & 18.8 & \textbf{5.9} \\
Full ReAct & 37.7 & 27.1 & 31.5 & 33.1 \\
\bottomrule
\end{tabular*}
\endgroup
\caption{Large-scale attribution diagnostic on a fixed 600-question subset of FRAMES shared across methods. CitePrec is the proportion of cited passages that entail their associated claims; CiteRec is the proportion of claims supported by at least one entailing citation; CiteF1 is their harmonic mean; Uncited is the proportion of claims with no citation. Metrics are aggregated over extracted claims and reported as percentages.}
\label{tab:attribution-results}
\end{table}

\textbf{Attribution quality.} Table~\ref{tab:attribution-results} compares three representative memory paradigms on the same 600 FRAMES questions: unbounded transcripts (Full-Trajectory ReAct), synthesized reports (ReportMemory), and structured trees (ReTree). FlatUpdate is omitted because it shares ReTree's atomic evidence and source indexing; Table~\ref{tab:main-results} instead uses it as the mechanism-matched control for downstream repair. ReTree attains the highest Citation Precision and Recall (44.5/41.3), compared with 37.7/27.1 for Full-Trajectory ReAct and 16.8/21.4 for ReportMemory. ReportMemory has the lowest Uncited Rate (5.9\%), but its cited passages are substantially less likely to support their associated claims. On this focused diagnostic, ReTree therefore provides more faithful claim-to-passage attribution than the full-trajectory and report-style baselines while maintaining a bounded per-step reasoning context.

\begin{figure*}[!t]
    \centering
    \includegraphics[width=\textwidth]{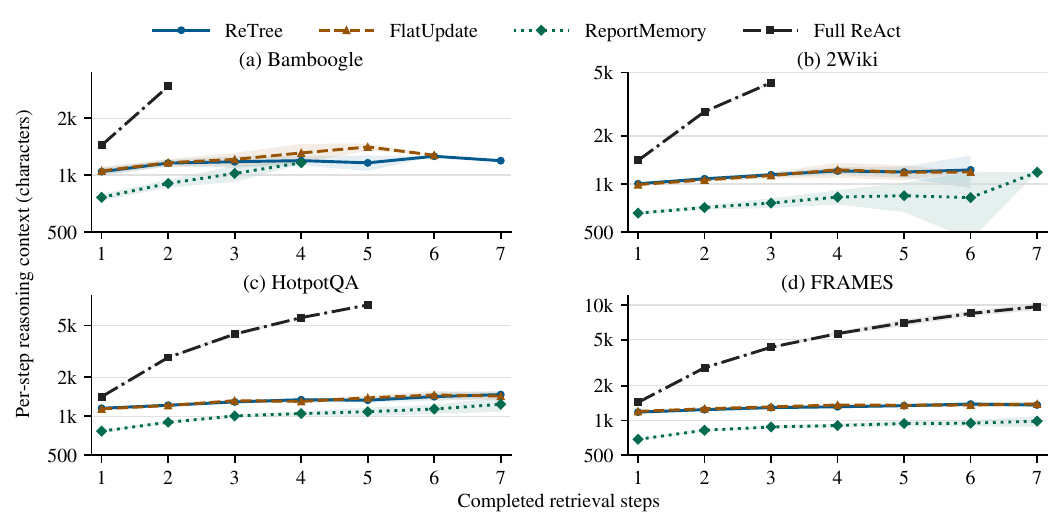}
    \caption{Variation in per-step reasoning context with retrieval step. At retrieval depth $t$, lines report the mean rendered reasoning context among runs that reach that depth; shading shows pointwise 95\% confidence intervals where estimable. Table~\ref{tab:main-results} instead averages each question's maximum context across steps. Panels use logarithmic y-axes with dataset-specific limits.}
    \label{fig:context-growth}
\end{figure*}

\textbf{Per-step context footprint and compute}. Table~\ref{tab:main-results} reports the average peak context load per run $M_q=\max_t c_{q,t}$. Across datasets, Full-Trajectory ReAct requires 1.27--1.51× as much peak context as ReTree. While ReportMemory achieves the smallest footprint (699--817 characters), its overall accuracy drops by 3.0 pp relative to ReTree, demonstrating that aggressive context stripping alone cannot preserve downstream reasoning.

Figure~\ref{fig:context-growth} traces how per-step policy context evolves across completed retrieval steps. Bounded memory variants remain stably constrained within a narrow band. In contrast, Full-Trajectory ReAct exhibits unbounded trajectory-dependent growth, surpassing 2.8K characters after two retrievals on 2Wiki and HotpotQA, and surpassing 5.6K after four retrievals on FRAMES. Crucially, natural backtracking is triggered in 9.6--17.5\% of ReTree runs, confirming that real-world retrieval actively exercises state repair. This structural repair incurs a modest overhead: ReTree uses 7--11\% more model calls and 10--13\% more total tokens than FlatUpdate.

\section{Limitations}

ReTree is a working-memory mechanism for search rather than a general long-term memory. Its hard-pruning policy treats every descendant of a revised node as dependent, although some downstream facts may remain valid. An incorrect conflict decision can therefore remove useful state. Future work could replace hard subtree pruning with finer-grained dependency graphs that track explicit claim-to-evidence and summary-to-evidence links, enabling more selective repair.

Bounding the single-step context window does not eliminate external compute and memory overheads. As retrieval progresses, the external evidence store grows continuously, making linear evidence selection and tree-traversal search operations slower over very long horizons. Furthermore, structural repair—including conflict confirmation, summary regeneration, and search resumption—requires additional language model queries, making ReTree use 7--11\% more model calls and 10--13\% more tokens than FlatUpdate. Vector indexing, cached retrievals, batched operations, and specialized lightweight conflict detectors could substantially mitigate these computational costs in future deployments.

\section{Conclusion}

We introduced ReTree, a tree-structured working memory that treats retrieval-induced error cascades as a structural state-repair problem. ReTree combines a bounded per-step reasoning context, stable claim-to-source provenance, and contradiction-triggered backtracking that repairs an introducing node before invalidating dependent descendants. Across 2,149 questions drawn from four public benchmarks, ReTree improves judge accuracy over Full-Trajectory ReAct by 8.3--25.6 points; the average maximum per-step reasoning context of Full-Trajectory ReAct is $1.27$--$1.51\times$ that of ReTree. ReTree also improves accuracy over FlatUpdate on every dataset and exceeds ReportMemory, the strongest aggregate baseline, by 3.0 points overall. On a large-scale FRAMES attribution diagnostic, ReTree also attains higher citation precision and recall than both the full-trajectory and report-style baselines. More broadly, working memory for search agents should not only compress context, but also preserve a revisable dependency structure over evidence and intermediate conclusions. In this view, memory is not merely a shorter transcript, but an editable state representation for provenance-aware repair.

\bibliographystyle{plainnat}
\bibliography{main}

\end{document}